\documentclass[10pt,twocolumn,letterpaper]{article}

\usepackage{cvpr}      

\usepackage[dvipsnames, table]{xcolor}

\definecolor{mygray}{gray}{0.9}
\definecolor{sparklinecolor}{named}{red}
\usepackage{sparklines}

\definecolor{cvprblue}{rgb}{0.21,0.49,0.74}

\usepackage[pagebackref,breaklinks,colorlinks,citecolor=cvprblue]{hyperref}

\def\paperID{7} 
\def\confName{CVPR4Animal}
\def\confYear{2024}

\title{Markerless retro-identification complements re-identification of individual insect subjects in archived image data of biological experiments}

\author{Asaduz Zaman\\
Dept. of Data Science and Artificial Intelligence\\
Faculty of Information Technology,\\
Monash University, Australia\\
{\tt\small asaduzzaman@monash.edu}
\and
Vanessa Kellermann\\
Dept. of Environment and Genetics\\
School of Agriculture, Biomedicine, and Environment,\\
La Trobe University, Australia\\
{\tt\small v.kellermann@latrobe.edu.au}
\and
Alan Dorin\\
Dept. of Data Science and Artificial Intelligence\\
Faculty of Information Technology,\\
Monash University, Australia\\
{\tt\small alan.dorin@monash.edu}
}

\begin{document}
\maketitle

\begin{abstract}
This study introduces markerless retro-identification of animals, a novel concept and practical technique to identify past occurrences of organisms in archived data, that complements traditional forward-looking chronological re-identification methods in longitudinal behavioural research. Identification of a key individual among multiple subjects may occur late in an experiment if it reveals itself through interesting behaviour after a period of undifferentiated performance. Often, longitudinal studies also encounter subject attrition during experiments. Effort invested in training software models to recognise and track such individuals is wasted if they fail to complete the experiment. Ideally, we would be able to select individuals who both complete an experiment and/or differentiate themselves via interesting behaviour, prior to investing computational resources in training image classification software to recognise them. We propose retro-identification for model training to achieve this aim. This reduces manual annotation effort and computational resources by identifying subjects only after they differentiate themselves late, or at an experiment's conclusion. Our study dataset comprises observations made of morphologically similar reed bees (\textit{Exoneura robusta}) over five days. We evaluated model performance by training on final day five data, testing on the sequence of preceding days, and comparing results to the usual chronological evaluation from day one. Results indicate no significant accuracy difference between models. This underscores retro-identification's value in improving resource efficiency in longitudinal animal studies.

\end{abstract}

\section{Introduction}
\label{sec:intro}
\begin{figure}
    \centering
    \includegraphics[width=\linewidth]{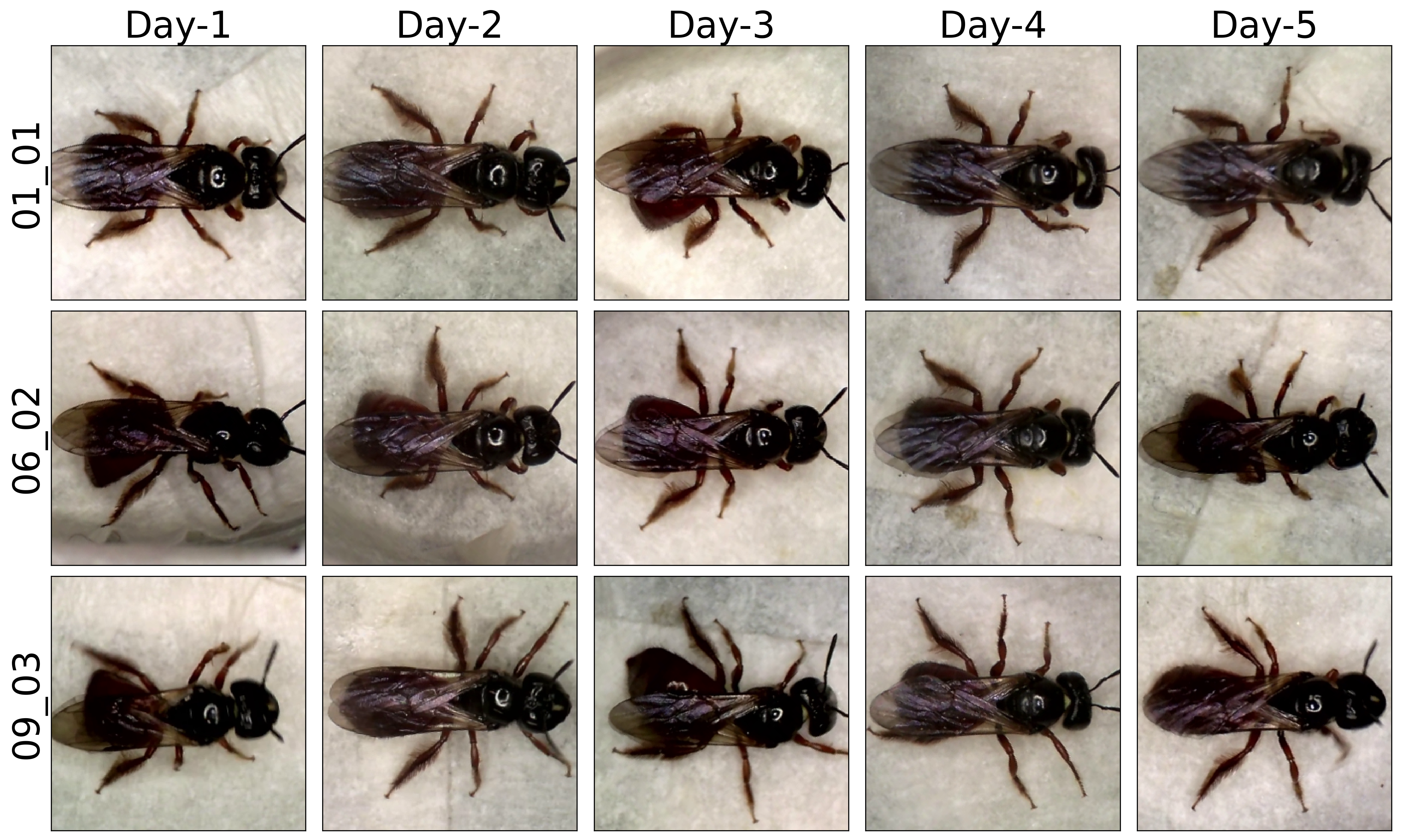}
    \caption{Sample images from our dataset: three reed bee specimens (rows) recorded over five days (columns).}
    \label{fig:random-bee-images}
\end{figure}
In longitudinal behavioural studies, tracking individual subjects over time, identifying them when they first appear, and again when they re-appear in subsequent observations, is critical for understanding behaviour \citep{Boenisch.2018}. Re-identification (re-id) of small, visually similar animals, such as honeybees, can be supported by physical markers or tags \citep{Mersch.2013, Crall.2015, Boenisch.2018, Meyers.2023}. However, these can alter subjects' behaviour \citep{Dennis.2008}. Markerless re-id potentially enables researchers to assess study subjects' natural behaviours \citep{Murali.2019}. However, this is difficult for highly similar individuals, such as insects, and requires algorithms to be trained, often on hand-annotated images \citep{Tausch.2020}. In experiments with insects, subject attrition through death or disappearance can be high during longitudinal studies over several days. This is especially true outside controlled lab settings, where the additional issue of morphological change through wear and tear may confound efforts to re-id an individual. If a subject is lost or visually altered during an experiment, resources invested in training image classification software to recognise it will potentially be wasted. This inefficiency is worsened by the need to conduct experiments on multiple subjects in the expectation that few will survive to the end, and of those, even fewer will exhibit a particular behaviour of interest, such as learning a task, solving a puzzle or collecting a specific resource \citep{Dona.2022}. Hence, late identification of key subjects from an initial larger starting set is common in longitudinal behavioural insect studies. How can researchers avoid wasted manual image annotation and re-id model training on subjects that do not ultimately contribute useful data? Here we propose and test \emph{retro}-identification (retro-id) to tackle this issue.

Rather than follow the convention of training models on initial (day one) data and attempting to follow individuals chronologically during an experiment, we propose it can sometimes be more useful to do the reverse. That is, sometimes we should train our algorithms on late-stage experimental image data of just the key (surviving or otherwise interesting) individuals. And then we should track these key individuals retrospectively through archived image data to explore their behaviour during the experiment. This focuses attention on annotation and model training for subjects critical to a study, rather than wasting resources on subjects that may not persist or exhibit relevant behaviour.

We hypothesise that a model trained on insect image data from day one and tested for its ability to re-id insects through to day N, would exhibit the same performance as a model trained on day N data and tested to retro-id insects back to day one. We tested this by monitoring 15 individual reed bees over 5 days. These semi-social pollinators have high phenotypic similarity (\Cref{fig:random-bee-images}) and are naturally found close to one another, even sharing nests, making re-id ecologically valuable but challenging. We trained several transfer learning-based image classification models using data from days 1 and 5, evaluating their accuracy on subsequent and preceding day sequences respectively. Below we review related work (\Cref{sec:related-works}), describe our data collection and model evaluation methods (\Cref{sec:method}), discuss results (\Cref{sec:results}) and conclude (\Cref{sec:conclusion}).

\section{Related Works}
\label{sec:related-works}
Explicit recognition of retro-id's value as distinct from re-id, and a need to test its performance are, to the best of our knowledge, novel. Re-id however, is well researched for human faces \citep{Taigman.2014, Liao.2015, Schroff.2015, Lisanti.2015, Wu.2016}, and somewhat so for insects \citep{Kastbeeger.2003, Mersch.2013, Crall.2015, Boenisch.2018, Murali.2019, Borlinghaus.2023, Meyers.2023}. Insect re-id algorithms may rely on small markers or tags attached to an insect to track it over separate observations \citep{Mersch.2013, Crall.2015, Boenisch.2018, Meyers.2023}. Six ant colonies were monitored using tags over 41 days, collecting approximately nine million social interactions to understand their behaviour \citep{Mersch.2013}. BEETag, a tracking system using bar codes, was used for automated honeybee tracking \citep{Crall.2015}, and \citet{Boenisch.2018} developed a QR-code system for honeybee lifetime tracking. \citet{Meyers.2023} demonstrated automated honeybee re-id by marking their thoraxes with paint, while demonstrating the potential of markerless re-id using their unmarked abdomens. Markerless re-id has been little explored. The study of Giant honeybees' wing patterns using size-independent characteristics and a self-organising map was a pioneering effort in non-invasive re-id \citep{Kastbeeger.2003}. Convolutional neural networks have been used for markerless fruit fly re-id \citep{Murali.2019} and triplet-loss-based similarity learning approaches have also been used to re-id Bumble bees returning to their nests \citep{Borlinghaus.2023}. 

All these studies adopt chronological re-id despite many highly relevant scenarios where this is inefficient. Our study therefore explores retro-id as a novel complementary approach to tracking individual insects for ecological and biological research.

\section{Method}
\label{sec:method}
\subsection{Data Collection}
\label{subsec:data-collection}
We obtained reed bees from the Dandenong Ranges National Park, Victoria, Australia (lat. -37.90, long. 145.37)\footnote{Parks Victoria permit number AA-0000010}. These bees exhibit semi-social behaviour and construct their nests within the pithy stems of fern fronds and other plants \citep{Cronin.1999}. Each nest can consist of several females who share brood-rearing and defence responsibilities. We placed each insect in a separate container to facilitate individual id for testing. In order to run the experiment over several days, insects were refrigerated overnight below 4°C. After warming up, each bee was individually recorded daily in an arena. Here it was illuminated by an overhead ring light and videoed using a Dino-Lite digital microscope for 30–50 seconds per session at 30 fps. We followed the process listed below to create our final datasets.

\begin{enumerate}
    \item Video Processing: Bee videos were processed frame by frame. To automate this, we trained a YOLO-v8 model to detect a bee's entire body, head, and abdomen in each frame. This enabled automatic establishment of the bee's orientation in the frame.
    
    \item Image Preparation: Upon detection, bees were cropped from the frames using the coordinates provided by Step 1. To align bees, we rotated frames using a bee's orientation before cropping. Centred on the detected entire bee body, a 400x400 pixel region (determined empirically for our bee/microscope setup) was cropped, then resized to 256x256.
    
    \item Contrast Adjustment: To enhance image quality and ensure uniform visibility across all samples, Contrast Limited Adaptive Histogram Equalisation (CLAHE) \citep{Pizer.1987} was applied.
    
    \item Quality Control: Manual inspection to remove misidentified objects maintained dataset integrity and ensured only bee images were included.
    
    \item Dataset Segregation: The final dataset was divided into image subsets, each from a single session, to avoid temporal data leakage.
\end{enumerate}

Using Steps 1–5, we curated a dataset of daily bee recording sessions across five consecutive days. Each session included the same 15 individuals videoed for approximately 1200 images/session (total dataset approximately 90K images).

\subsection{Network Architecture, Training, Evaluation}
We used a transfer-learning-based approach for re-/retro-id of the reed bees. All models were pre-trained on the ImageNet dataset \cite{Deng.2009} and subsequently fine-tuned using our own dataset. To identify suitable transfer-learning models, we selected 17 different models distributed across 10 different model architectures and parameter numbers ranging from $49.7$ million in $swin\_v2\_s$ to 0.73 million parameters in $squeezenet1\_0$. To evaluate the models, we collected a second set of data on Day 5, ``set-2'', four hours from the first set using Steps 1–5 (above). We trained all 17 models on the first set of Day 5 data. The 17 models were then evaluated based on their ability to re-id individuals in Day 5 set-2 data. From them, we selected the seven models with the highest Accuracy (and F1) scores for further consideration. We then trained this top-7 on our original Day 1 and Day 5 data. We evaluated Day 1 models forward on Day 2–5 data and Day 5 models back in time on Day 4–1 data to conduct our main experiments. These forward and backwards evaluations allowed comparison of markerless re- and retro- id of individual insects. The training process was similar for all of the models we considered. We have used Adam Optimiser with a learning rate of 0.001 with 0.0001 weight decay, with a total 100 epochs on the training dataset. We used cross-entropy loss as the loss function for these models. 

\section{Results and Discussion}
\label{sec:results}

\begin{table}[ht]
\centering
\begin{tabular}{lrrrrr}
\toprule
\textbf{Model Name}        & \textbf{Params} & \textbf{Accuracy} $\uparrow$ & \textbf{F1 Score} \\ 
\midrule
regnet\_y\_3\_2gf          & 17.95                   & \textbf{0.5342}    & \textbf{0.4791}   \\ \hline
shufflenet\_v2\_x2\_0     & 5.38                    & 0.4769             & 0.4311            \\ \hline
mnasnet0\_75              & 1.91                    & 0.4651             & 0.4180            \\ \hline
densenet121               & 6.97                    & 0.4618             & 0.3832            \\ \hline
googlenet                 & 5.62                    & 0.4448             & 0.3821            \\ \hline
mnasnet1\_3               & 5.02                    & 0.4428             & 0.3667            \\ \hline
regnet\_y\_400mf         & 3.91                    & 0.4216             & 0.3679            \\ \hline \rowcolor{mygray}
densenet201               & 18.12                   & 0.4044             & 0.3472            \\ \hline \rowcolor{mygray}
efficientnet\_v2\_s       & 20.20                   & 0.3953             & 0.3576            \\ \hline \rowcolor{mygray}
resnet18                  & 11.18                   & 0.3937             & 0.3299            \\ \hline \rowcolor{mygray}
shufflenet\_v2\_x1\_0     & 1.27                    & 0.3584             & 0.3028            \\ \hline \rowcolor{mygray}
mobilenet\_v3\_small      & 4.22                    & 0.3524             & 0.3572            \\ \hline \rowcolor{mygray}
efficientnet\_b0          & 4.03                    & 0.3343             & 0.3072            \\ \hline \rowcolor{mygray}
resnext50\_32x4d          & 23.01                   & 0.3303             & 0.2972            \\ \hline \rowcolor{mygray}
mobilenet\_v3\_large      & 1.53                   & 0.3210             & 0.2732            \\ \hline \rowcolor{mygray}
swin\_v2\_s               & \textbf{48.98}          & 0.2928             & 0.2440            \\ \hline \rowcolor{mygray}
squeezenet1\_1            & 0.73                    & 0.2670            & 0.2082            \\ \bottomrule
\end{tabular}
\caption{Performance metrics of pretrained models on Day 5 set-1 vs. set-2 datasets sorted by Accuracy. Shaded rows show models failing our performance threshold for further consideration. Bold numbers represent either the largest model or best-scores.}
\label{tab:model_performance_morning_vs_afternoon}
\end{table}


\begin{table*}[ht]
\centering
    \begin{tabular}{l|cccc|cccc|c}
    \toprule
    &  \multicolumn{4}{c}{Re-Identification}&  \multicolumn{4}{c}{Retro-Identification}&\\
    \midrule
    Model Name              &  Day 2        &  Day 3        &  Day 4        &  Day 5  &  Day 4     &  Day 3&  Day 2& Day 1 & Sparklines\\
    \midrule
    $densenet121$           &  0.33&  0.26&  0.24&  0.17&0.33& 0.22& 0.23& 0.10&
    \begin{sparkline}{10}
        \definecolor{sparklinecolor}{named}{blue}
        \spark 0 1.00 0.33 0.70 0.66 0.61 1 0.31 /
        \definecolor{sparklinecolor}{named}{orange}
        \spark 0 0.98 0.33 0.54 0.66 0.54 1 0.00 /
    \end{sparkline}
    \\
    $googlenet$             &  0.26&  0.23&  0.19&  0.21&0.32& 0.20& 0.14& 0.11&
    \begin{sparkline}{10}
        \definecolor{sparklinecolor}{named}{blue}
        \spark 0 0.70 0.33 0.55 0.66 0.37 1 0.45 /
        \definecolor{sparklinecolor}{named}{orange}
        \spark 0 1.00 0.33 0.42 0.66 0.10 1 0.00 /
    \end{sparkline}
    \\
    $mnasnet0\_75$          &  0.25&  0.29&  0.15&  0.17&0.24& 0.15& 0.28& 0.08&
    \begin{sparkline}{10}
        \definecolor{sparklinecolor}{named}{blue}
        \spark 0 0.80 0.33 1.00 0.66 0.32 1 0.44 /
        \definecolor{sparklinecolor}{named}{orange}
        \spark 0 0.77 0.33 0.32 0.66 0.99 1 0.00 /
    \end{sparkline}
    \\
    $mnasnet1\_3$           &  0.37&  0.35&  \textbf{0.35}&  0.15&\textbf{0.35}&  0.26&  0.26& 0.17&
    \begin{sparkline}{10}
        \definecolor{sparklinecolor}{named}{blue}
        \spark 0 1.00 0.33 0.88 0.66 0.90 1 0.00 /
        \definecolor{sparklinecolor}{named}{orange}
        \spark 0 0.88 0.33 0.53 0.66 0.51 1 0.12 /
    \end{sparkline}
    \\
    $regnet\_y\_400mf$   & 0.36& 0.37& 0.23& 0.22&0.28& 0.29& 0.23&0.24&
    \begin{sparkline}{10}
        \definecolor{sparklinecolor}{named}{blue}
        \spark 0 0.91 0.33 1.00 0.66 0.05 1 0.00 /
        \definecolor{sparklinecolor}{named}{orange}
        \spark 0 0.39 0.33 0.44 0.66 0.05 1 0.11 /
    \end{sparkline}
    \\
    $regnet\_y\_3\_2gf$      & 0.36& \textbf{0.39}& 0.30& \textbf{0.30}&\textbf{0.35}& \textbf{0.32}& \textbf{0.38}&\textbf{0.34}&
    \begin{sparkline}{10}
        \definecolor{sparklinecolor}{named}{blue}
        \spark 0 0.64 0.33 1.00 0.66 0.05 1 0.00 /
        \definecolor{sparklinecolor}{named}{orange}
        \spark 0 0.61 0.33 0.24 0.66 0.86 1 0.42 /
    \end{sparkline}
    \\
    $shufflenet\_v2\_x2\_0$ & \textbf{0.39}& \textbf{0.39}& 0.25& 0.22&0.29& 0.27& 0.29&0.17&
    \begin{sparkline}{10}
        \definecolor{sparklinecolor}{named}{blue}
        \spark 0 0.98 0.33 1.00 0.66 0.36 1 0.20 /
        \definecolor{sparklinecolor}{named}{orange}
        \spark 0 0.53 0.33 0.43 0.66 0.53 1 0.00 /
    \end{sparkline}
    \\ 
 \midrule
    Mean & $0.33$& $0.33$& $0.24$& $0.20$&$0.31$& $0.24$& $0.26$&$0.17$&
    \begin{sparkline}{10}
        \definecolor{sparklinecolor}{named}{blue}
        \spark 0 1.00 0.33 0.97 0.66 0.45 1 0.20 /
        \definecolor{sparklinecolor}{named}{orange}
        \spark 0 0.86 0.33 0.45 0.66 0.54 1 0.00 /
    \end{sparkline}
    \\
    $\pm$ std.dev& $\pm 0.06$& $\pm 0.07$& $\pm 0.07$& $\pm 0.05$&$\pm 0.04$& $\pm 0.06$& $\pm 0.07$&$\pm 0.09$&\\
 \bottomrule
    \end{tabular}
\caption{Re-identification and retro-identification accuracy of selected models. Models trained on day 1 and tested on days 2–5 chronologically are labelled as Re-Identification, and models trained on day 5 and tested on days 4-1 backwards are labelled as Retro-Identification. Re-id and retro-id accuracies are plotted as blue and orange sparklines respectively. Bold numbers represent best accuracy for that specific dataset.}
\label{tab:model_performance_all_datasets_all}
\end{table*}

\begin{figure}
    \centering
    \includegraphics[width=\linewidth]{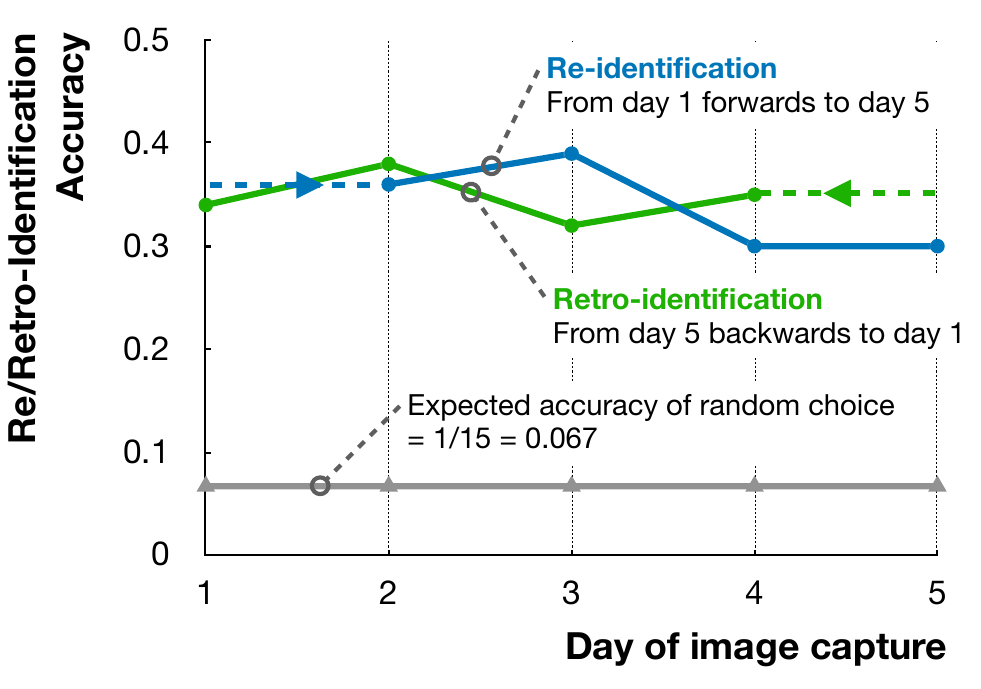}
    \caption{Re/retro-identification accuracy of $regnet\_y\_3\_2gf$ model where re-identification is shown as forward identification from day 1-5, and retro-identification is shown as backward identification from day 5-1.}
    \label{fig:day-offset}
\end{figure}

Our initial model selection (\Cref{tab:model_performance_morning_vs_afternoon}), demonstrates these models' ability to re-id reed bees over a short 4 hour period. For comprehensive analysis using all datasets, we chose models with Accuracy $>0.40$ (and F1 score $>0.35$), seven in total (unshaded in \Cref{tab:model_performance_morning_vs_afternoon}). Notably, the $mnasnet0\_75$, a very small network with only 1.91M parameters, outperformed many of its larger counterparts. This suggests that smaller models that are practical for their training and inference speed may be useful and underscores the need to consider a range of network architectures for suitability in specific domains. The $regnet$ and $mnasnet$ architectures demonstrated superior performance for our problem.

The comparative assessment of re- and retro-id across our five day experiment for all seven models appears in \Cref{tab:model_performance_all_datasets_all} while \Cref{fig:day-offset} demonstrates re- and retro-id accuracy of the top performing model ($regnet\_y\_3\_2\_gf$). We found a decline in day-to-day re-id accuracy for all models, consistent with findings from \citep{Murali.2019}. A similar trend was observed in retro-id for almost all of the models, as demonstrated by the sparkline graphs in the \Cref{tab:model_performance_all_datasets_all}. Additionally, a two-tailed t-test of the mean accuracies across the days revealed no significant difference in the trends of re- and retro-id average accuracies (t-Stat=$0.725$, p=$0.495$) for our reed bee dataset, thus confirming our initial hypothesis. 

Retro-id of animal subjects \emph{after the fact}, and tracking their appearances back through an experiment, can offer a more efficient approach than the conventional chronological re-id. This might be the case when many insects fail to complete a study or when only a few exhibit a target behaviour. For instance, if we were monitoring honeybee hives for bees returning with a specific pollen colour \citep{Ngo.2021} indicative of an invasive weed \citep{Gillbank.2013, Batchelor.2023}, not many arrivals would be expected to meet our criteria from the thousands of departures each day. Yet if a bee does arrive meeting our criterion, we might like to establish when she departed to estimate her travel time/distance, or even identify from the dance she watched on the honeycomb which direction she is likely to have flown \citep{Wario.2015, Wario.2017}. Trying to id every insect of the large number that leaves the hive, then re-id them again as they return is infeasible and unnecessary. Our retro-id approach would be suitable in this application. Similarly, reed bees share nests and interact semi-socially, making individual identification in natural settings over long periods of time difficult. Using retro-id we only need to annotate data and train algorithms to recognise reed bees that last the full length of an experiment, which would in turn allow us to track complex behaviour and help us understand how often and for how long reed bees forage.

Combining re-id and retro-id provides flexibility to conduct analyses in both temporal directions. Since accuracy decreases gradually as we move away from the training day, if it's possible to identify an individual of interest midway through an experiment, it would also be advantageous to train models on data from that day. We could then apply both retro- and re-id to potentially reduce the impacts of poor identification performance at temporal extremes. 

Domain adaptation \citep{Ganin.2016} is another interesting approach for re-identification that treats changes in appearance over a long period of time as a domain shift. Networks that can find and learn domain-invariant features can be trained as effective re-identification models \citep{Murali.2019}, an approach worth exploring in the future for retro-identification as well. 

Overall, our results suggest that retro-identification is a viable and valuable alternative to re-id for longitudinal studies. It is especially helpful when there is an expectation of high subject attrition or late identification of key subjects. Retro-identification is resource-efficient as it allows users to conduct data annotation and model training only on individual experimental subjects of known interest.

\section{Conclusion}
\label{sec:conclusion}
In this study, we examined the applicability of retro-identification, a conceptual approach and practical, simple technique that identifies past occurrences of experimental subjects in archival image data, in contrast to previous forward-looking re-identification approaches. To demonstrate the approach, we developed a challenging reed bee re-identification dataset of 15 individuals recorded over 5 days. From a pool of 17 transfer learning-based image classification models, 7 were selected for analysis based on their performance on training and test datasets captured over a short 4-hour time frame. Subsequently, we trained the top-7 models on Day 1 data, and evaluated their performance on long time frame data in the form of sequences of subsequent data from Day 2 to Day 5 data. And we did the reverse, training the models on Day 5 data and testing our approach by retro-identifying individuals back through the preceding days of data in the archive to Day 1. We found no statistical evidence of differences between re- and retro- identification accuracy, underscoring the potential of retro-identification in longitudinal studies to help manage the sometimes infeasible demands of forward re-identification on large numbers of experimental test subjects, many of whom may ultimately not contribute any useful data to an experimental outcome. 

{
    \small
    \bibliographystyle{ieeenat_fullname}
    \bibliography{body}

\begin{thebibliography}{24}
\providecommand{\natexlab}[1]{#1}
\providecommand{\url}[1]{\texttt{#1}}
\expandafter\ifx\csname urlstyle\endcsname\relax
  \providecommand{\doi}[1]{doi: #1}\else
  \providecommand{\doi}{doi: \begingroup \urlstyle{rm}\Url}\fi

\bibitem[Batchelor et~al.(2023)Batchelor, Bell, Campos, and Webber]{Batchelor.2023}
Kathryn~L. Batchelor, Karen~L. Bell, Mariana Campos, and Bruce~L. Webber.
\newblock {Can honey bees be used to detect rare plants? Taking an eDNA approach to find the last plants in a weed eradication program}.
\newblock \emph{Environmental DNA}, 5\penalty0 (6):\penalty0 1516--1526, 2023.

\bibitem[Boenisch et~al.(2018)Boenisch, Rosemann, Wild, Dormagen, Wario, and Landgraf]{Boenisch.2018}
Franziska Boenisch, Benjamin Rosemann, Benjamin Wild, David Dormagen, Fernando Wario, and Tim Landgraf.
\newblock {Tracking All Members of a Honey Bee Colony Over Their Lifetime Using Learned Models of Correspondence}.
\newblock \emph{Frontiers in Robotics and AI}, 5:\penalty0 35, 2018.

\bibitem[Borlinghaus et~al.(2023)Borlinghaus, Tausch, and Rettenberger]{Borlinghaus.2023}
Parzival Borlinghaus, Frederic Tausch, and Luca Rettenberger.
\newblock {A Purely Visual Re-ID Approach for Bumblebees (Bombus terrestris)}.
\newblock \emph{Smart Agricultural Technology}, 3:\penalty0 100135, 2023.

\bibitem[Crall et~al.(2015)Crall, Gravish, Mountcastle, and Combes]{Crall.2015}
James~D. Crall, Nick Gravish, Andrew~M. Mountcastle, and Stacey~A. Combes.
\newblock {BEEtag: A Low-Cost, Image-Based Tracking System for the Study of Animal Behavior and Locomotion}.
\newblock \emph{PLoS ONE}, 10\penalty0 (9):\penalty0 e0136487, 2015.

\bibitem[Cronin and Schwarz(1999)]{Cronin.1999}
Adam~L. Cronin and Michael~P. Schwarz.
\newblock {Life Cycle and Social Behavior in a Heathland Population of Exoneura robusta (Hymenoptera: Apidae): Habitat Influences Opportunities for Sib Rearing in a Primitively Social Bee}.
\newblock \emph{Annals of the Entomological Society of America}, 92\penalty0 (5):\penalty0 707--716, 1999.

\bibitem[Deng et~al.(2009)Deng, Dong, Socher, Li, Li, and Fei-Fei]{Deng.2009}
Jia Deng, Wei Dong, Richard Socher, Li-Jia Li, Kai Li, and Li Fei-Fei.
\newblock {ImageNet: A large-scale hierarchical image database}.
\newblock \emph{2009 IEEE Conference on Computer Vision and Pattern Recognition}, pages 248--255, 2009.

\bibitem[Dennis et~al.(2008)Dennis, Newberry, Cheng, and Estevez]{Dennis.2008}
R.~L. Dennis, R.~C. Newberry, H.-W. Cheng, and I. Estevez.
\newblock {Appearance Matters: Artificial Marking Alters Aggression and Stress}.
\newblock \emph{Poultry Science}, 87\penalty0 (10):\penalty0 1939--1946, 2008.

\bibitem[Dona et~al.(2022)Dona, Solvi, Kowalewska, Mäkelä, MaBouDi, and Chittka]{Dona.2022}
Hiruni Samadi~Galpayage Dona, Cwyn Solvi, Amelia Kowalewska, Kaarle Mäkelä, HaDi MaBouDi, and Lars Chittka.
\newblock {Do bumble bees play?}
\newblock \emph{Animal Behaviour}, 194:\penalty0 239--251, 2022.

\bibitem[Ganin et~al.(2016)Ganin, Ustinova, Ajakan, Germain, Larochelle, Laviolette, March, and Lempitsky]{Ganin.2016}
Yaroslav Ganin, Evgeniya Ustinova, Hana Ajakan, Pascal Germain, Hugo Larochelle, François Laviolette, Mario March, and Victor Lempitsky.
\newblock {Domain-Adversarial Training of Neural Networks}.
\newblock \emph{Journal of Machine Learning Research}, 17\penalty0 (59):\penalty0 1--35, 2016.

\bibitem[Gillbank(2013)]{Gillbank.2013}
Linden Gillbank.
\newblock Hawkweeds: a recent discovery in {Victoria}’s {Alps} and a taxonomic name change, 2013.
\newblock Accessed: 2024-05-22, 
\newblock https://invasives.org.au/blog/hawkweeds-a-recent-discovery-in-victorias-alps-and-a-taxonomic-name-change/.

\bibitem[Kastberger et~al.(2003)Kastberger, Radloff, and Kranner]{Kastbeeger.2003}
Gerald Kastberger, Sarah Radloff, and Gerhard Kranner.
\newblock {Individuality of wing patterning in Giant honey bees (Apis laboriosa)}.
\newblock \emph{Apidologie}, 34\penalty0 (3):\penalty0 311--318, 2003.

\bibitem[Liao et~al.(2015)Liao, Hu, Zhu, and Li]{Liao.2015}
Shengcai Liao, Yang Hu, Xiangyu Zhu, and Stan~Z. Li.
\newblock {Person Re-Identification by Local Maximal Occurrence Representation and Metric Learning}.
\newblock \emph{2015 IEEE Conference on Computer Vision and Pattern Recognition (CVPR)}, pages 2197--2206, 2015.

\bibitem[Lisanti et~al.(2015)Lisanti, Masi, Bagdanov, and Bimbo]{Lisanti.2015}
Giuseppe Lisanti, Iacopo Masi, Andrew~D. Bagdanov, and Alberto~Del Bimbo.
\newblock {Person Re-Identification by Iterative Re-Weighted Sparse Ranking}.
\newblock \emph{IEEE Transactions on Pattern Analysis and Machine Intelligence}, 37\penalty0 (8):\penalty0 1629--1642, 2015.

\bibitem[Mersch et~al.(2013)Mersch, Crespi, and Keller]{Mersch.2013}
Danielle~P. Mersch, Alessandro Crespi, and Laurent Keller.
\newblock {Tracking Individuals Shows Spatial Fidelity Is a Key Regulator of Ant Social Organization}.
\newblock \emph{Science}, 340\penalty0 (6136):\penalty0 1090--1093, 2013.

\bibitem[Meyers et~al.(2023)Meyers, Cordero, Bravo, Noel, Agosto-Rivera, Giray, and Mégret]{Meyers.2023}
Luke Meyers, Josué~Rodríguez Cordero, Carlos~Corrada Bravo, Fanfan Noel, José Agosto-Rivera, Tugrul Giray, and Rémi Mégret.
\newblock {Towards Automatic Honey Bee Flower-Patch Assays with Paint Marking Re-Identification}.
\newblock \emph{arXiv}, 2023.

\bibitem[Murali et~al.(2019)Murali, Schneider, Levine, and Taylor]{Murali.2019}
Nihal Murali, Jonathan Schneider, Joel~D. Levine, and Graham~W. Taylor.
\newblock {Classification and Re-Identification of Fruit Fly Individuals Across Days with Convolutional Neural Networks}.
\newblock \emph{2019 IEEE Winter Conference on Applications of Computer Vision (WACV)}, 00:\penalty0 570--578, 2019.

\bibitem[Ngo et~al.(2021)Ngo, Rustia, Yang, and Lin]{Ngo.2021}
Thi~Nha Ngo, Dan Jeric~Arcega Rustia, En-Cheng Yang, and Ta-Te Lin.
\newblock {Automated monitoring and analyses of honey bee pollen foraging behavior using a deep learning-based imaging system}.
\newblock \emph{Computers and Electronics in Agriculture}, 187:\penalty0 106239, 2021.

\bibitem[Pizer et~al.(1987)Pizer, Amburn, Austin, Cromartie, Geselowitz, Greer, Romeny, Zimmerman, and Zuiderveld]{Pizer.1987}
Stephen~M. Pizer, E.~Philip Amburn, John~D. Austin, Robert Cromartie, Ari Geselowitz, Trey Greer, Bart ter~Haar Romeny, John~B. Zimmerman, and Karel Zuiderveld.
\newblock {Adaptive histogram equalization and its variations}.
\newblock \emph{Computer Vision, Graphics, and Image Processing}, 39\penalty0 (3):\penalty0 355--368, 1987.

\bibitem[Schroff et~al.(2015)Schroff, Kalenictbhenko, and Philbin]{Schroff.2015}
Florian Schroff, Dmitry Kalenictbhenko, and J.ames Philbin.
\newblock {FaceNet: A Unified Embedding for Face Recognition and Clustering}.
\newblock \emph{2015 IEEE Conference on Computer Vision and Pattern Recognition (CVPR)}, pages 815--823, 2015.

\bibitem[Taigman et~al.(2014)Taigman, Yang, Ranzato, and Wolf]{Taigman.2014}
Yaniv Taigman, Ming Yang, Marc'~Aurelio Ranzato, and Lior Wolf.
\newblock {DeepFace: Closing the Gap to Human-Level Performance in Face Verification}.
\newblock \emph{2014 IEEE Conference on Computer Vision and Pattern Recognition}, pages 1701--1708, 2014.

\bibitem[Tausch et~al.(2020)Tausch, Stock, Fricke, and Klein]{Tausch.2020}
Frederic Tausch, Simon Stock, Julian Fricke, and Olaf Klein.
\newblock {Bumblebee Re-Identification Dataset}.
\newblock \emph{2020 IEEE Winter Applications of Computer Vision Workshops (WACVW)}, 00:\penalty0 35--37, 2020.

\bibitem[Wario et~al.(2015)Wario, Wild, Couvillon, Rojas, and Landgraf]{Wario.2015}
Fernando Wario, Benjamin Wild, Margaret~J. Couvillon, Raúl Rojas, and Tim Landgraf.
\newblock {Automatic methods for long-term tracking and the detection and decoding of communication dances in honeybees}.
\newblock \emph{Frontiers in Ecology and Evolution}, 3:\penalty0 103, 2015.

\bibitem[Wario et~al.(2017)Wario, Wild, Rojas, and Landgraf]{Wario.2017}
Fernando Wario, Benjamin Wild, Raúl Rojas, and Tim Landgraf.
\newblock {Automatic detection and decoding of honey bee waggle dances}.
\newblock \emph{PLOS ONE}, 12\penalty0 (12):\penalty0 e0188626, 2017.

\bibitem[Wu et~al.(2016)Wu, Chen, Li, Wu, You, and Zheng]{Wu.2016}
Shangxuan Wu, Ying-Cong Chen, Xiang Li, An-Cong Wu, Jin-Jie You, and Wei-Shi Zheng.
\newblock {An Enhanced Deep Feature Representation for Person Re-Identification}.
\newblock \emph{2016 IEEE Winter Conference on Applications of Computer Vision (WACV)}, pages 1--8, 2016.

\end{thebibliography}
    
}

\end{document}